\documentclass[conference]{IEEEtran}

\IEEEoverridecommandlockouts
\usepackage{cite}
\usepackage{amsmath,amssymb,amsfonts}
\usepackage{algorithm}
\usepackage{algpseudocode}
\usepackage{graphicx}
\usepackage{hyperref}
\usepackage{textcomp}
\usepackage{soul}
\usepackage{caption}
\usepackage{subcaption}
\usepackage[table,xcdraw]{xcolor}
\def\BibTeX{{\rm B\kern-.05em{\sc i\kern-.025em b}\kern-.08em
    T\kern-.1667em\lower.7ex\hbox{E}\kern-.125emX}}

\makeatletter
\def\ps@IEEEtitlepagestyle{%
	\def\@oddfoot{\mycopyrightnotice}%
	\def\@evenfoot{}%
}
\def\mycopyrightnotice{%
	{\hfill \footnotesize 978-1-6654-5705-7/23/\$31.00~\copyright 2023 IEEE\hfill}
}
\makeatother

\begin{document}

\title{Intelligent Client Selection for Federated Learning using Cellular Automata

}

\makeatletter
\newcommand{\linebreakand}{%
  \end{@IEEEauthorhalign}
  \hfill\mbox{}\par
  \mbox{}\hfill\begin{@IEEEauthorhalign}
}
\makeatother

\author{
    \IEEEauthorblockN{
       Nikolaos Pavlidis\IEEEauthorrefmark{1},
       Vasileios Perifanis\IEEEauthorrefmark{1},
       Theodoros Panagiotis Chatzinikolaou\IEEEauthorrefmark{1}, \\
       Georgios Ch. Sirakoulis\IEEEauthorrefmark{1} and
       Pavlos S. Efraimidis\IEEEauthorrefmark{1}, 
    }
    \IEEEauthorblockA{
        \IEEEauthorrefmark{1} Department of Electrical and Computer Engineering, Democritus University of Thrace, Xanthi, Greece \\
        }
    \IEEEauthorblockA{e-mail: npavlidi@ee.duth.gr}
    
}


\newcommand{\vasilis}[1]{\textcolor{red}{ #1}}
\newcommand{\nikos}[2]{\textcolor{blue}{ #1}}

\maketitle
\thispagestyle{plain}
\pagestyle{plain}

\begin{abstract}
Federated Learning (FL) has emerged as a promising solution for privacy-enhancement and latency minimization in various real-world applications, such as transportation, communications, and healthcare. FL endeavors to bring Machine Learning (ML) down to the edge by harnessing data from million of devices and IoT sensors, thus enabling rapid responses to dynamic environments and yielding highly personalized results. However, the increased amount of sensors across diverse applications poses challenges in terms of communication and resource allocation, hindering the participation of all devices in the federated process and prompting the need for effective FL client selection. To address this issue, we propose Cellular Automaton-based Client Selection (CA-CS), a novel client selection algorithm, which leverages Cellular Automata (CA) as models to effectively capture spatio-temporal changes in a fast-evolving environment. CA-CS considers the computational resources and communication capacity of each participating client, while also accounting for inter-client interactions between neighbors during the client selection process, enabling intelligent client selection for online FL processes on data streams that closely resemble real-world scenarios. In this paper, we present a thorough evaluation of the proposed CA-CS algorithm using MNIST and CIFAR-10 datasets, while making a direct comparison against a uniformly random client selection scheme. Our results demonstrate that CA-CS achieves comparable accuracy to the random selection approach, while effectively avoiding high-latency clients.
\end{abstract}


\section{Introduction}
Federated Learning (FL)~\cite{McMahan2017FedAvg} has gained substantial attention in current research, standing out as one of the most popular machine learning topics regarding privacy concerns and latency minimization. Its wide-ranging applications in various domains like autonomous driving~\cite{elbir2022fedvehicular}, recommendation systems~\cite{Perifanis2022FedNCF, Perifanis2023FedPOIRec}, communications~\cite{Perifanis20225g} and healthcare~\cite{Joshi2022FedHealth} have established it as a comprehensive solution for privacy-enhanced, low-latency distributed machine learning. The market value for federated learning is expected to double by 2030 \cite{PolarisReport}, providing a significant impetus for the development of new algorithms aimed at enhancing FL efficiency and performance.

Despite the remarkable progress in FL algorithms and its diverse applications, several limitations and unresolved issues still remain. Two of the most significant constraints of FL involve the presence of non-identically and independently distributed \textit{(non-iid)} data due to the massive distribution of data among clients~\cite{Xiaodong2022noniid}, which limits its predictive accuracy and the increased \textit{communications costs} brought on by the random selection process during federated rounds~\cite{bano2022fedtcs}. A promising solution to address these issues is to perform intelligent client selection methods. For instance, Lei Fu et al.~\cite{LeiFu2022ClientSelection} propose that an effective client selection scheme could serve as a solution to address data heterogeneity while simultaneously enhancing fairness, robustness, model accuracy, and reducing communication costs.

In the context of FL, data is decentralized across multiple data silos owned by different clients. Each client conducts local training using its private data, resulting in local model parameters. Only the learnable parameters, typically model weights, are transmitted to a central server. The central server then aggregates these received model parameters to create a global model, which is transmitted back to each client for updating their local parameters. FL clients are often numerous, comprising millions of smart devices or IoT sensors globally. However, due to potential connection issues or limited processing capacity, not all clients may be consistently available to participate in the FL process.
To avoid heavy computational and communication overhead associated with training and propagating weights across millions of devices simultaneously, FL comprises a client selection mechanism~\cite{Lo2022Patterns}. This technique involves selecting only a subset of clients to participate in each federated round. Although this strategy alleviates some of the computational burden, it may introduce fairness and robustness issues, leading to decreased accuracy, particularly when the data distribution among clients is non-iid.

A cellular automaton (CA)~\cite{Wolfram1983} refers to a discrete computational model comprising a grid of cells, where each cell exists in one of a finite set of states. These cells' states evolve over time based on a predefined set of rules, which rely on the states of their neighboring cells. Complex systems can be investigated through CA, where computational power is merged with mathematical and physical foundations \cite{von1966theory}, with examples to be evident in epidemic propagation \cite{sirakoulis2000}, chemical computing representation \cite{tsompanas2022}, and resolution of complex logic puzzles \cite{chatzinikolaou2022}. In the context of client selection in FL, a cellular grid can be utilized to represent clients, where each cell is associated with a fitness value denoting the suitability or eligibility of the respective client for participation in the FL process. During each federated round, the cell states undergo updating procedures, taking into account the states of neighboring cells and adhering to specific updating rules. This mechanism allows the CA to perform computations, resembling a client selection scheme for FL.

Motivated by the above, we present a novel client selection algorithm for FL that leverages CA. This algorithm takes into account both the computational capabilities and communication limitations of each participating entity, as well as the standard deviation of their data. The proposed approach is thoroughly evaluated across multiple FL scenarios, and the results demonstrate its potential, as it identifies and excludes stragglers while maintaining high prediction accuracy.

In this paper, we introduce a novel federated client selection algorithm by leveraging CA to capture clients' computational and communication capabilities, and we formulate an experimental setting to comprehensively evaluate our proposed algorithm in a vehicular network problem using the CIFAR-10 and MNIST datasets. In this direction, we pave the way for investigating inter-client relations during client selection, which to the best of our knowledge, has not been adequately addressed in the related literature.

The rest of this paper is structured as follows: Section \ref{related_work} introduces the main concepts used for effective client selection in FL and reviews the related work. Section \ref{methodology} defines the settings of our experimental scenario inspired by vehicular networks as well as the equations describing the CA used for client selection. Section \ref{experiments} discusses the experimental results. Finally, Section \ref{conclusion} concludes our work and proposes further research directions.

\section{Related Work}
\label{related_work}

Several works address the topic of client selection in FL aiming to improve both performance of the predictive models based on data and model quality, as well as efficiency of the FL process by minimizing training and communication times and taking into consideration resource capabilities of each client.

Multiple papers have focused on biased client selection strategies to improve the error convergence speed and efficiency in FL. Cho et al. \cite{cho2020banditbased} propose the UCB-CS strategy, which leverages bandit-based techniques to achieve faster convergence with lower communication overhead, thereby addressing intermittent client availability and communication constraints.
Moreover, the issue of optimizing communication efficiency between the server and clients has been addressed through innovative sampling approaches. Fraboni et al. \cite{fraboni2021clustered} introduced a clustered sampling approach, which leads to better client representativeness, reduced variance in the clients local weights, and improved training convergence and variability compared to standard sampling approaches.
Similarly, Shen et al. \cite{shen2022fast} presented a clustering-based approach to accelerate the convergence of FL by reducing variance and generating client subsets with certain representativeness of sampling.
Huang at al. \cite{huang2022stochastic} addressed the issue of volatile clients, who may experience training failures due to various reasons. They proposed a stochastic client selection scheme, E3CS, that considers effective participation and fairness and achieves faster convergence while maintaining high accuracy.
In the context of FL in vehicular networks, where heterogeneity among network users is a significant factor, Cha et al. \cite{Cha2022Fuzzy} introduced a fuzzy logic-based client selection scheme. This approach considers various factors, such as the number of local samples, sample freshness, computation capability, and available network throughput, to choose appropriate clients for participation in the training process. In this paper, we emphasize on exploring the inter-client relations among immediate neighboring nodes. These connections frequently emerge as a result of local network topologies, geographical arrangements, or logical associations. To this scope, we leverage a Cellular Automaton adept at capturing these types of relationships for effective FL client selection.

\section{Methodology}
\label{methodology}

In this section, we formulate the problem for client selection in FL and we propose a novel client selection algorithm based on a CA.

\begin{figure}
    \centering
    \includegraphics[width=\columnwidth]{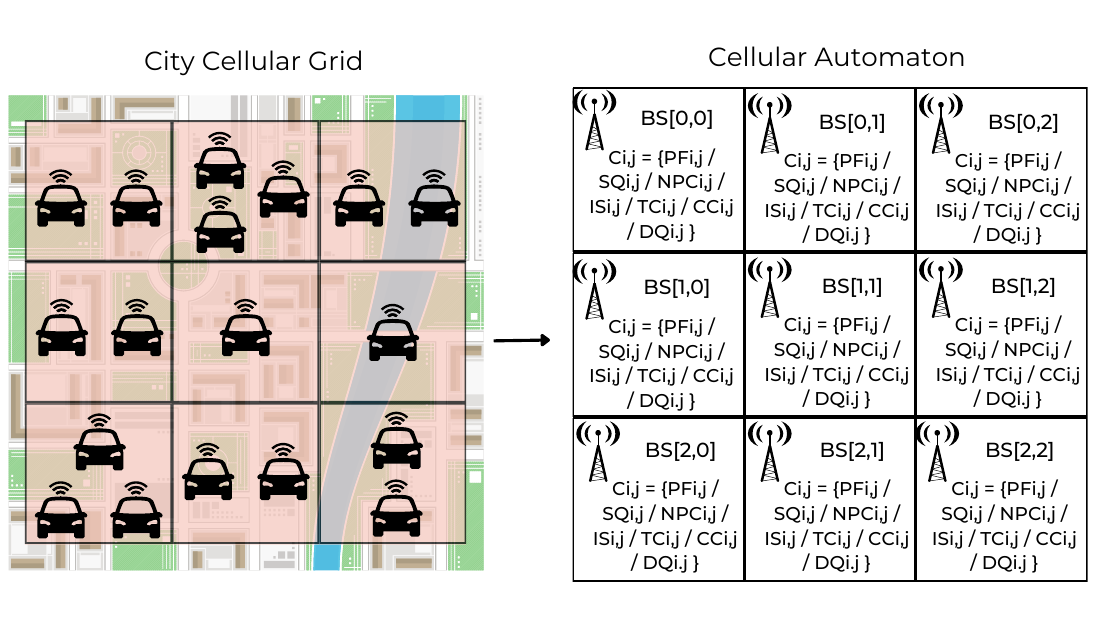}
    \caption{Experiment Architecture and CA analogy.}
    \label{fig:experiment_arch}
\end{figure}

\subsection{Problem Statement and FL Formulation}

We consider a city that is divided in an $M\times M$ cellular grid, as shown in Fig.~\ref{fig:experiment_arch}. In total, there are $K = M^2$ cells, each one has each own base station that is capable to receive and send messages and perform training of machine learning models. These base stations are the participating clients of the federated process. We also consider $N$ vehicles moving around the city, carrying their own private data and being assigned to their closest base station. At the first round, each participating vehicle $n \in N$ starts with an initial set of data, denoted by $x_0^n$ and is assigned to its closest base station. At every timestep $t$, a random subset $S \in N$ of vehicles moves to a different cell. When a vehicle is assigned to another base station it gets an additional set of data that is being added to it's private dataset, i.e., $X_t^n = \bigcup_{i=0}^{t} X_t^n$. For each FL node, an indicator, called Throughput Congestion (TC), is calculated based mainly on the samples quantity of the respective node and it's participating one hop neighbors. We assume that this is a metric indicating the communication capabilities of the node, with higher TC meaning more traffic around the node, thus decreased communication throughput available. Whenever, a client from the top 20\% of those with the highest TC is selected, the whole simulation is delayed by 5 seconds, denoting a penalty for selecting stragglers. The percentage of clients that are considered as stragglers and the associated time penalty are indicative and related to the specific problem formulation.

\begin{figure}[b]
    \centering
    \includegraphics[width=\columnwidth]{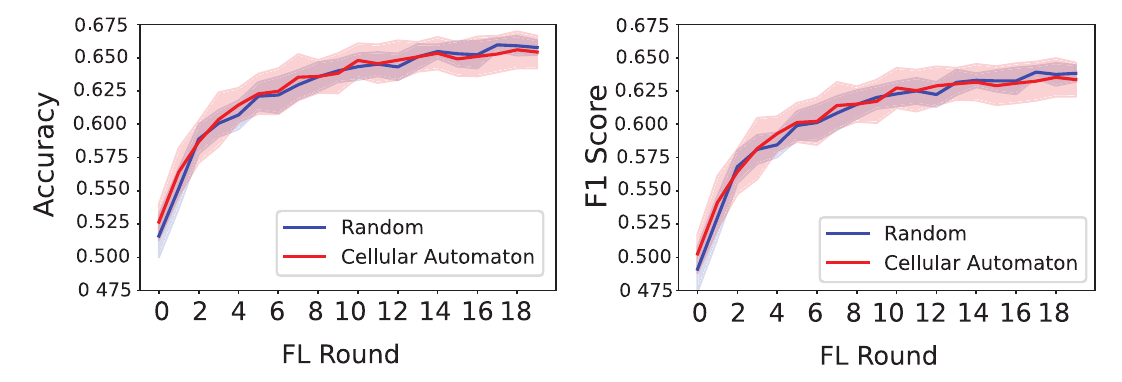}
    \caption{CIFAR10 - Accuracy \& F1 Score}
    \label{fig:cifar_acc_f1}
\end{figure}

\subsection{CA Equations and Client Selection Algorithm}

The assumptions of the problem formulated in the previous paragraph, create the demand for a selection algorithm that can leverage information from neighboring nodes and thus avoid selecting nodes with high TC. For this reason, a CA is considered, with each of its cells representing an FL client.

Each of the cells has the following state:

$C^t_{i,j} = \{ PF^t_{i,j}, SQ^t_{i,j}, NPC^t_{i,j}, IS^t_{i,j}, TC^t_{i,j}, CC^t_{i,j}, DQ^t_{i,j}\},$
where:

\begin{itemize} 
    \item $PF$: A participation flag that denotes the participation or not of the FL node in the FL process based on previous output of the algorithm.
    \item $SQ$: Sample Quantity, based on the number of vehicles that are registered to each cell's base station.
    \item $NPC$: A Non Participation Counter, that is computed based on previous outputs and showing how many FL rounds the node has not been selected.
    \item $IS$: Inbound samples, based on the number of vehicles registered to a base station.
    \item $TC$: Throughput Congestion, showing the congestion on the available bandwidth based on neighbors participation and samples quantity.
    \item $CC$: Computational Capacity, computed by multiplying SQ with a random variable $d$ showing possible computational problems.
    \item $DQ$: Distribution Quality, computed at each timestep as the standard deviation of the samples.
\end{itemize}

The output of each cell is computed using the following equation:

$C^t_{i,j} = (\alpha \cdot \frac{IS^t_{i,j} }{SQ^t_{i,j}}+ \beta \cdot NPC^t_{i,j} + \gamma \cdot \frac{1}{DQ^t_{i,j})}) \cdot \frac{CC^t_{i,j}} {TC^t_{i,j}}$

where $\alpha, \beta,\gamma$ are weights used for normalization (typically set to 0.33 if no normalization is needed). $SQ$, $IS$ and $DQ$ are measured at each timestep by the respective base station and considered as input to the CA.

The updating rules for other CA's parameters are:
\begin{itemize}
    \item $NPC^{t+1}_{i,j} = 0$  if  $PF^t_{i,j} = 1$  else  $NPC^{t+1}_{i,j} = NPC^{t}_{i,j} + 1$
    \item $CC^{t+1}_{i,j} = d^{t+1} \cdot SQ^{t+1}_{i,j} $ where $d$ is a uniform random variable $\in [0,1]$
    \item $TC^{t+1}_{i,j} = PF^{t}_{i,j} \cdot SQ^{t}_{i,j} + e \cdot [PF^{t}_{i+1,j} \cdot SQ^{t}_{i+1,j} +  PF^{t}_{i-1,j} \cdot SQ^{t}_{i-1,j} +  PF^{t}_{i,j+1} \cdot SQ^{t}_{i,j+1} +  PF^{t}_{i,j-1} \cdot SQ^{t}_{i,j-1}] + m \cdot [PF^{t}_{i+1,j+1} \cdot SQ^{t}_{i+1,j+1} +  PF^{t}_{i-1,j-1} \cdot SQ^{t}_{i-1,j-1} +  PF^{t}_{i-1,j+1} \cdot SQ^{t}_{i-1,j+1} +  PF^{t}_{i+1,j-1} \cdot SQ^{t}_{i+1,j-1}]$ where $e$, $m$ are constants that denote the importance of Von Neumann's neighborhood over Moore's.
\end{itemize}

After computing the output $C^t_{i,j}$ of each cell, we select the $c\%$ with the highest output to participate in the FL process, and we change their $PF$ respectively (1 if selected, 0 otherwise). The parameter $c$ denotes the percentage of the participating clients and is set to $40\%$ as a demonstrative value for the conducted experiments. Determining the value of parameter $c$ is a challenging task that involves dynamically balancing the exploration and exploitation aspects and is also influenced by other unknown factors, such as connectivity issues.

    
    




\begin{figure}[b]
    \centering
    \includegraphics[width=\columnwidth]{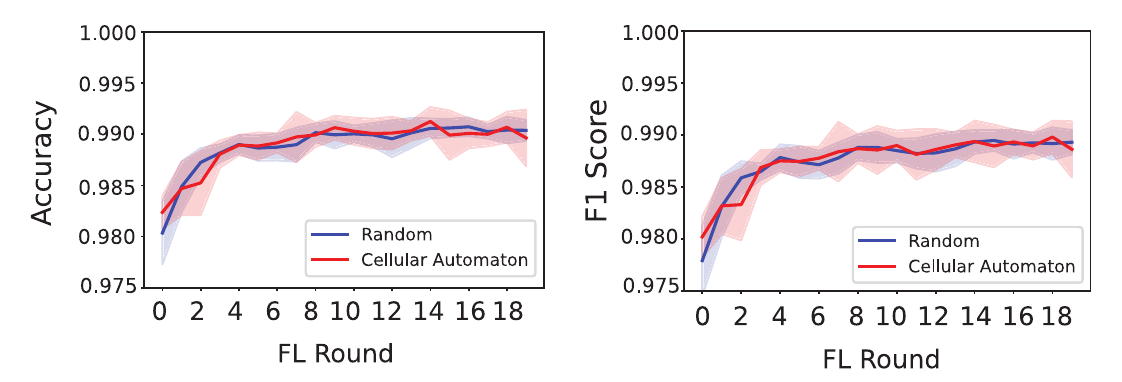}
    \caption{MNIST - Accuracy \& F1 Score}
    \label{fig:mnist_acc_f1}
\end{figure}

\section{Experiments}
\label{experiments}
In this section, we outline the experimental setup and present the results, focusing on the predictive accuracy and execution time.
The experiments were conducted on a workstation running Windows 11 with 16 GB memory and AMD Ryzen 5 4600H CPU. This setup closely mirrors real-world scenarios involving clients with moderate computational capabilities. To ensure the validity of the results, 10 different seeds were used for the initialization of random generators.\footnote{Code available at \url{https://github.com/nikopavl4/CA\_Client\_Selection/}.}

\begin{figure}[t]
    \centering
    \includegraphics[width=\columnwidth]{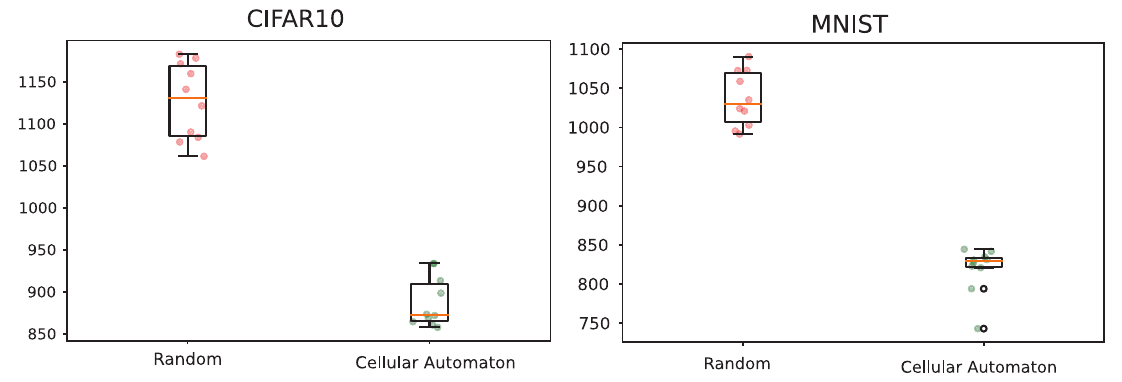}
    \caption{Execution Time for different client selection algorithms}
    \label{fig:execution_times}
\end{figure}

\subsection{Experimental Settings}
In order to effectively resemble a real-world machine learning problem for vehicles, we utilized two widely used image datasets, MNIST and CIFAR-10, with 70.000 and 60.000 samples respectively. An 80-20\% train-test split was performed, with the training set being dispensed to 100 vehicles and the test set being stored at the central server. A $5\times5$ CA was considered, resembling the 25 participating clients.

\subsection{Learning Settings}
For both classification tasks, we consider a Convolutional Neural Network (CNN), exactly as described in FL original paper by McMahan et al.~\cite{McMahan2017FedAvg} with two $5\times5$ convolution layers, a fully connected layer with 512 units and ReLu activation, and a final softmax output layer, alongside the Adam optimizer and 0.001 learning rate. The learning process was repeated for 20 FL rounds under the FedAvg aggregation algorithms~\cite{McMahan2017FedAvg} and 5 local epochs per client, while Accuracy and F1 Score were utilized to evaluate the performance of the ML model.

\subsection{Results}

In this section, we present the results of our experiments comparing the novel Cellular Automata-based Client Selection (CA-CS) algorithm with a uniformly random sampling algorithm in the context of federated learning.

Figs.~\ref{fig:cifar_acc_f1} and \ref{fig:mnist_acc_f1} present the accuracy and F1 scores obtained for the CIFAR-10 and MNIST datasets, respectively, using both random selection and the CA-CS algorithm. As can be observed, the CA-CS algorithm demonstrated similar results to the random selection in terms of predictive accuracy and F1 score both for CIFAR-10 and MNIST datasets.

Fig.~\ref{fig:execution_times} demonstrates the execution times for different experiments using both random and CA-CS client selection algorithms for the CIFAR-10 and MNIST datasets. Notably, the CA-CS algorithm significantly outperformed the random selection approach, leading to as much as 25\% lower execution times. This finding underscores the effectiveness of the CA in avoiding clients with high-latency, indicated by high Throughput Congestion (TC) values during specific rounds.

The results suggest that the CA-CS algorithm successfully achieves its aim when utilized as a client selection scheme, based on the parameters and targets set by the designer. However, it's essential to recognize that the update rule we designed for the CA in our example imposed execution time penalties on clients with high numbers of samples, i.e., those with substantial communication traffic. Additionally, clients with one-hop neighbors with high sample quantities were also affected by these penalties, assuming that excessive communication traffic in one neighborhood might lead to delays in other nodes within the network topology.

\section{Conclusions and Future Work}
\label{conclusion}

In this paper, we proposed a novel approach to client selection in FL using CA. Our CA-CS algorithm intelligently captures clients' computational and communication capabilities, enhancing the efficiency of FL systems. Through comprehensive experiments with CIFAR-10 and MNIST datasets, we evaluated the CA-CS algorithm, comparing it to random selection. While the CA-CS algorithm did not significantly outperform random selection in terms of predictive accuracy and F1 scores, it excelled in reducing execution times by avoiding high-latency clients, making it a valuable asset in resource-constrained scenarios. The evaluation of the CA-CS performance in systems with uniform levels of congestion will need to be carried out through future experiments.

Overall, our research demonstrates the potential of CA for effective client selection in FL, setting the stage for further advancements and optimizations in this area for real-world applications.
Possible future directions would include but are not limited to modifying the CA update rule to address other critical challenges, such as data and model heterogeneity, fairness, communication costs, and resource allocation, especially when these factors heavily rely on the data within one-hop neighborhoods. Finally, other spatio-temporal models, e.g. graphs, could be utilized to capture inter-client relations.



\bibliographystyle{IEEEtran}
\bibliography{main}

\begin{thebibliography}{10}
\providecommand{\url}[1]{#1}
\csname url@samestyle\endcsname
\providecommand{\newblock}{\relax}
\providecommand{\bibinfo}[2]{#2}
\providecommand{\BIBentrySTDinterwordspacing}{\spaceskip=0pt\relax}
\providecommand{\BIBentryALTinterwordstretchfactor}{4}
\providecommand{\BIBentryALTinterwordspacing}{\spaceskip=\fontdimen2\font plus
\BIBentryALTinterwordstretchfactor\fontdimen3\font minus
  \fontdimen4\font\relax}
\providecommand{\BIBforeignlanguage}[2]{{%
\expandafter\ifx\csname l@#1\endcsname\relax
\typeout{** WARNING: IEEEtran.bst: No hyphenation pattern has been}%
\typeout{** loaded for the language `#1'. Using the pattern for}%
\typeout{** the default language instead.}%
\else
\language=\csname l@#1\endcsname
\fi
#2}}
\providecommand{\BIBdecl}{\relax}
\BIBdecl

\bibitem{McMahan2017FedAvg}
B.~McMahan, E.~Moore, D.~Ramage, S.~Hampson, and B.~A.~y. Arcas,
  ``{Communication-Efficient Learning of Deep Networks from Decentralized
  Data},'' in \emph{Proceedings of the 20th AISTATS Conference}.

\bibitem{elbir2022fedvehicular}
A.~M. Elbir, B.~Soner, S.~Coleri, D.~Gunduz, and M.~Bennis, ``Federated
  learning in vehicular networks,'' 2022.

\bibitem{Perifanis2022FedNCF}
V.~Perifanis and P.~S. Efraimidis, ``Federated neural collaborative
  filtering,'' \emph{Knowledge-Based Systems}, vol. 242, p. 108441, 2022.

\bibitem{Perifanis2023FedPOIRec}
V.~Perifanis, G.~Drosatos, G.~Stamatelatos, and P.~S. Efraimidis, ``Fedpoirec:
  Privacy-preserving federated poi recommendation with social influence,''
  \emph{Information Sciences}, vol. 623, pp. 767--790, 2023.

\bibitem{Perifanis20225g}
V.~Perifanis, N.~Pavlidis, R.-A. Koutsiamanis, and P.~S. Efraimidis,
  ``Federated learning for 5g base station traffic forecasting,''
  \emph{Computer Networks}, p. 109950, 2023.

\bibitem{Joshi2022FedHealth}
M.~Joshi, A.~Pal, and M.~Sankarasubbu, ``Federated learning for healthcare
  domain - pipeline, applications and challenges,'' \emph{ACM Trans. Comput.
  Healthcare}, vol.~3, no.~4, nov 2022.

\bibitem{PolarisReport}
{POLARIS Market Research}, ``Federated learning market share, size, trends,
  industry analysis report.'' 2021.

\bibitem{Xiaodong2022noniid}
X.~Ma, J.~Zhu, Z.~Lin, S.~Chen, and Y.~Qin, ``A state-of-the-art survey on
  solving non-iid data in federated learning,'' \emph{Future Generation
  Computer Systems}, vol. 135, pp. 244--258, 2022.

\bibitem{bano2022fedtcs}
S.~Bano, N.~Tonellotto, P.~Cassar{\`a}, and A.~Gotta, ``Fedtcs: Federated
  learning with time-based client selection to optimize edge resources,'' 2022.

\bibitem{LeiFu2022ClientSelection}
L.~Fu, H.~Zhang, G.~Gao, H.~Wang, M.~Zhang, and X.~Liu, ``Client selection in
  federated learning: Principles, challenges, and opportunities,'' 2022.

\bibitem{Lo2022Patterns}
S.~K. Lo, Q.~Lu, L.~Zhu, H.-Y. Paik, X.~Xu, and C.~Wang, ``Architectural
  patterns for the design of federated learning systems,'' \emph{Journal of
  Systems and Software}, vol. 191, p. 111357, 2022.

\bibitem{Wolfram1983}
S.~Wolfram, ``Statistical mechanics of cellular automata,'' \emph{Reviews of
  Modern Physics}, vol.~55, no.~3, pp. 601--644, Jul. 1983.

\bibitem{von1966theory}
J.~V. Neumann, \emph{Theory of Self-Reproducing Automata}.\hskip 1em plus 0.5em
  minus 0.4em\relax University of Illinois Press, 1966.

\bibitem{sirakoulis2000}
G.~Sirakoulis, I.~Karafyllidis, and A.~Thanailakis, ``A cellular automaton
  model for the effects of population movement and vaccination on epidemic
  propagation,'' \emph{Ecological Modelling}, vol. 133, no.~3, pp. 209--223,
  2000.

\bibitem{tsompanas2022}
M.-A. Tsompanas, T.~P. Chatzinikolaou, and G.~C. Sirakoulis, ``Cellular
  automata application on chemical computing logic circuits,'' in
  \emph{Cellular Automata}.\hskip 1em plus 0.5em minus 0.4em\relax Cham:
  Springer International Publishing, 2022, pp. 3--14.

\bibitem{chatzinikolaou2022}
T.~P. Chatzinikolaou, R.-E. Karamani, and G.~C. Sirakoulis, ``Irregular
  learning cellular automata for the resolution of complex logic puzzles,''
  in \emph{Cellular Automata}.\hskip 1em plus 0.5em minus 0.4em\relax Springer
  International Publishing, 2022.

\bibitem{cho2020banditbased}
Y.~J. Cho, S.~Gupta, G.~Joshi, and O.~Yağan, ``Bandit-based
  communication-efficient client selection strategies for federated learning,''
  2020.

\bibitem{fraboni2021clustered}
Y.~Fraboni, R.~Vidal, L.~Kameni, and M.~Lorenzi, ``Clustered sampling:
  Low-variance and improved representativity for clients selection in federated
  learning,'' 2021.

\bibitem{shen2022fast}
G.~Shen, D.~Gao, D.~Song, L.~Yang, X.~Zhou, S.~Pan, W.~Lou, and F.~Zhou, ``Fast
  heterogeneous federated learning with hybrid client selection,'' 2022.

\bibitem{huang2022stochastic}
T.~Huang, W.~Lin, L.~Shen, K.~Li, and A.~Y. Zomaya, ``Stochastic client
  selection for federated learning with volatile clients,'' 2022.

\bibitem{Cha2022Fuzzy}
N.~Cha, Z.~Du, C.~Wu, T.~Yoshinaga, L.~Zhong, J.~Ma, F.~Liu, and Y.~Ji, ``Fuzzy
  logic based client selection for federated learning in vehicular networks,''
  \emph{IEEE Open Journal of the Computer Society}, vol.~3, pp. 39--50, 2022.

\end{thebibliography}


\end{document}